\documentclass{article}
\usepackage{arxiv}
\AtBeginDocument{\setlength{\headheight}{24pt}}
\usepackage[utf8]{inputenc}
\usepackage[T1]{fontenc}
\usepackage{hyperref}
\usepackage{url}
\usepackage{graphicx}
\usepackage{natbib}
\usepackage{algorithm}
\usepackage{algorithmic}
\usepackage{amsmath}
\usepackage{amssymb}
\usepackage{amsfonts}
\usepackage{amsthm}
\usepackage{mathtools}
\usepackage{booktabs}
\usepackage{multirow}
\usepackage{float}

\hypersetup{
  pdftitle={IAPO: Influence-Aware Policy Optimization for Credit Assignment in Multi-Turn Service Agents},
  pdfauthor={Bo Ren, Yirong Mao, Yi Yang, Wenhui Que},
  pdfkeywords={multi-turn agents, reinforcement learning, credit assignment, tool use}
}

\newtheorem{proposition}{Proposition}

\theoremstyle{definition}

\newcommand{\ms}[2]{#1\,$\pm$\,#2}
\newcommand{\errflag}{e}

\title{IAPO: Influence-Aware Policy Optimization \\ for Credit Assignment in Multi-Turn Service Agents}

\author{%
Bo Ren$^{\dagger}$,
Yirong Mao$^{\dagger}$,
Yi Yang,
Wenhui Que$^{*}$\\
WeChat, Tencent Inc.
}

\begin{document}

\maketitle
\footnotetext{$^{\dagger}$ Equal contribution; $^{*}$ Corresponding author.}

\begin{abstract}
Large Language Model (LLM) agents increasingly solve long-horizon tasks through multi-turn interactions with users and external tools. In these settings, relevant task information often unfolds over time rather than being fully specified at the initial prompt. Service agents make this challenge especially concrete: users may clarify or revise their goals, while tool responses provide information needed for subsequent decisions. Thus, a final reward alone cannot indicate which actions contributed to resolving the task. Recent methods rely on comparative evidence from other trajectories or resampled continuations, or on separately constructed step-level learning signals, to refine credit. However, a completed rollout already records how information and errors flow between agent actions. We introduce \textbf{Influence-Aware Policy Optimization (IAPO)}, which represents each rollout as a typed influence-dependency graph over trainable agent actions, with user and tool observations serving as evidence. IAPO converts support-use and failed-use structure into routing weights that redistribute the same trajectory-level advantage. Experiments with Qwen3-4B and Qwen3-8B demonstrate superior performance over multi-turn reinforcement learning (RL) baselines across three service-agent benchmarks: $\tau^2$-Bench, UserBench, and AgentChangeBench. BFCL-v4 Multi-Turn further shows that these gains do not compromise multi-turn function-calling performance. This work advances the understanding of credit assignment in multi-turn user interactions and provides a principled approach to training service agents from sparse outcome feedback.
\end{abstract}

\section{Introduction}
\label{sec:intro}

\begin{figure}[t]
\centering
\includegraphics[width=0.82\linewidth]{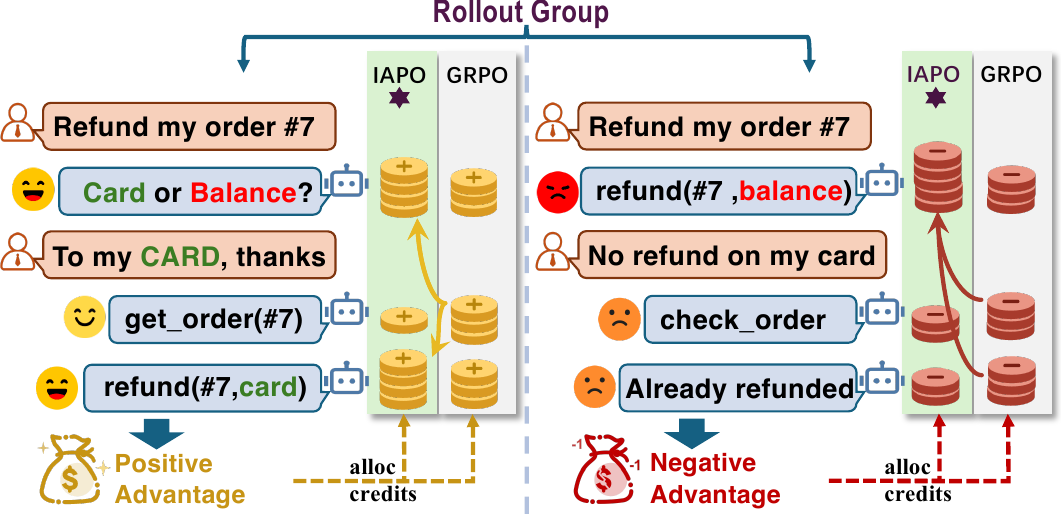}
\caption{Per-action advantage allocation for a successful (left, positive-advantage) and a failed (right, negative-advantage) refund rollout. Refunding to the original card or store balance is \textbf{correct} \textbf{only if it matches the user's clarified intent}, so an early clarifying turn decides whether later actions are right. GRPO applies the same advantage to every action, whereas IAPO routes credit according to realized support use and error propagation.}
\label{fig:teaser}
\end{figure}

Large Language Models (LLMs)~\citep{achiam2023gpt,team2023gemini,yang2025qwen3,liu2024deepseekv3} have moved from single-turn responders to interactive agents that plan, call tools, and act over multi-turn interactions with users and environments~\citep{yao2022react,schick2023toolformer,liu2024agentbench}. In particular, multi-turn service agents often solve tasks whose relevant information is not available at the initial prompt. Instead, users may reveal preferences or revise goals over successive turns, while tool calls expose operational states that determine which actions are valid or useful. For example, in an after-sales service task (Figure~\ref{fig:teaser}), before issuing the refund, the agent may need to clarify whether the user wants the money returned to the original card or converted into store balance, and then retrieve the order record needed to execute that choice. Thus, the contribution of an early action may become identifiable only after a later user clarification or tool response reveals how its information is used. We refer to this setting as \emph{evolving-state credit assignment}: task goals, constraints, and operational states are progressively instantiated through interaction rather than fully specified at the outset.

Group-based RL algorithms such as GRPO~\citep{shao2024deepseekmath} and its variants~\citep{yu2026dapo,zheng2025gspo} apply the same group-relative advantage to every trainable token of a sampled rollout. However, this trajectory-level signal does not distinguish how different intermediate actions contribute to the final outcome (Figure~\ref{fig:teaser}, GRPO columns). This limitation motivates finer-grained credit assignment for evolving-state interactions.

Existing approaches obtain finer-grained credit by supplementing the final outcome with additional evidence. One line compares recurring or matched states and resampled continuations across sampled trajectories~\citep{feng2026gigpo,ji2025treegrpo,samanta2026resets}. Another line constructs separate step-level signals through additional evaluation, using masked-context policy scoring, a teacher model, or a learned critic on training-time information~\citep{kong2026infopo,xie2026tips,zhou2025sweetrl,tan2026hcapo}.

However, obtaining this finer credit adds a requirement that the final outcome does not. 1) Cross-rollout methods need comparable structure across sampled trajectories. GiGPO~\citep{feng2026gigpo} relies on environment states that recur across rollouts, and branch- or reset-based methods~\citep{ji2025treegrpo,samanta2026resets} rely on a forkable or resettable environment with additional continuation samples. Free-form user and tool dialogue rarely provides either, because distinct user turns and tool returns leave few comparable state groups and a live user cannot be faithfully reset or branched. 2) Separately-scored methods add a scoring channel outside the reward, such as masked-context policy scoring~\citep{kong2026infopo}, a teacher model~\citep{xie2026tips}, or a learned critic on privileged information~\citep{zhou2025sweetrl}. The resulting step signal is computed apart from task completion and can drift from it. No existing method redistributes outcome credit using only the dependency structure already realized within a single completed interaction.

In this work, we observe that a completed user--tool rollout already records more structure than its outcome reward exposes: a later action may consume information elicited or supplied by an earlier one, or reuse an earlier invalid output and fall back because of an observable error. Therefore, we propose \textbf{IAPO}, which represents each completed rollout as a typed influence-dependency graph over trainable agent actions and routes the original trajectory-level advantage according to observed support-use and failed-use dependencies, using user and tool observations as evidence without assigning them policy-gradient credit. Since this dependency structure is already realized in the transcript and tool log, it can redistribute the same trajectory-level advantage without changing the reward, rollout process, group normalization, or clipped-loss implementation. IAPO keeps the critic-free, group-based update intact while introducing finer, action-level credit. The change lies in which action tokens carry the feedback, not in the value of the final task outcome.

\paragraph{Contributions.}
\begin{itemize}
  \item We identify realized downstream information use as a credit object for evolving-state service interactions.
  \item We introduce IAPO, which extracts typed within-rollout dependencies and routes the trajectory-level advantage through support-use and failed-use structure.
  \item We prove mass conservation, sign preservation, and uniform fallback, and show that IAPO leads on the three service-agent benchmarks across two Qwen3 scales and maintains competitive out-of-distribution retention on BFCL-v4 Multi-Turn.
\end{itemize}

\begin{figure*}[t]
\centering
\includegraphics[width=0.90\textwidth]{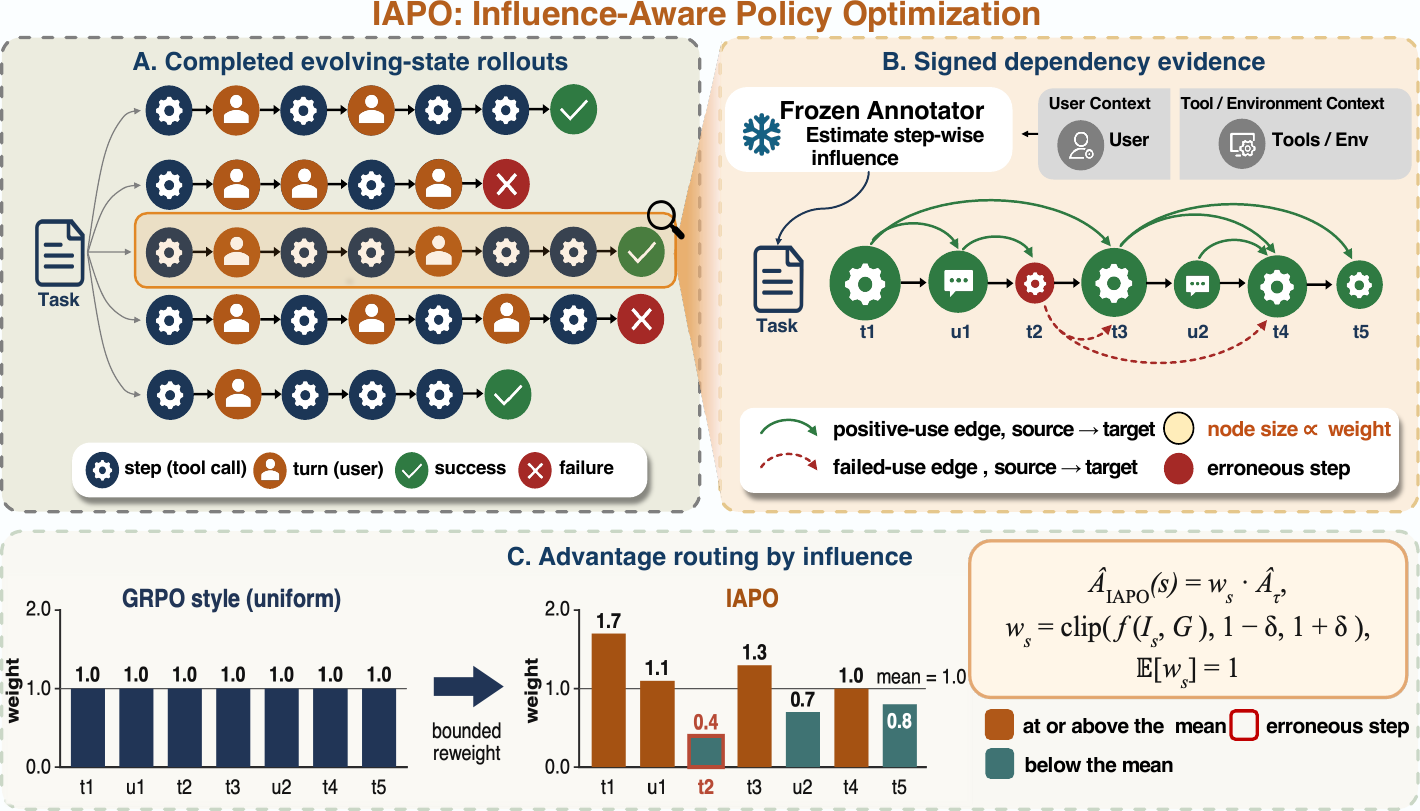}
\caption{Overview of IAPO. (A) Multiple rollouts are sampled per task and scored by the outcome reward. (B) A frozen annotator extracts signed influence edges among trainable actions, using user and tool observations as evidence. (C) The resulting graph features are normalized into bounded per-action weights that replace GRPO's uniform advantage with sign-conditioned routing.}
\label{fig:principle}
\end{figure*}

\section{Related Work}
\label{sec:related}

\paragraph{Interactive benchmarks for user-interacting agents.}
Task-oriented dialogue has long been treated as sequential decision making~\citep{budzianowski2018multiwoz,ultes2017pydial}. Interactive benchmarks have progressively incorporated simulated users, stateful tools, and multi-turn feedback~\citep{yao2024taubench,barres2025tau2bench,qian2025userbench,wang2023mint}. Beyond dialogue, agent benchmarks cover web navigation, embodied environments, and API-based tool use~\citep{yao2022webshop,shridhar2021alfworld,li2023apibank,qin2024toolllm,zhou2024webarena,deng2023mind2web,liu2024agentbench,mialon2023gaia}. More recent testbeds further expose incremental preference acquisition and explicit goal shifts across interaction episodes~\citep{qian2025userbench,rana2025agentchangebench}. These settings expose evolving user goals and tool-mediated state, but leave open how a final outcome should be attributed to the actions that gathered, used, or corrupted intermediate evidence.

\paragraph{RL for LLMs and credit assignment.}
Early LLM agents used frozen models with prompting and tool use, such as ReAct~\citep{yao2022react} and Toolformer~\citep{schick2023toolformer}; more recent work optimizes agents directly with reinforcement learning, building on PPO~\citep{schulman2017ppo}, GRPO~\citep{shao2024deepseekmath}, and their variants~\citep{zheng2025gspo,yu2026dapo,liu2025drgrpo}. Since outcome-only rewards leave intermediate steps unattributed, a growing body of work assigns credit more finely, from classical return decomposition and process supervision~\citep{arjonamedina2019rudder,harutyunyan2019hindsight,lightman2024verify,cui2025prime,ng1999rewardshaping} to recent agent-oriented methods. On reasoning and agent tasks, GiGPO~\citep{feng2026gigpo} groups recurring states across rollouts, tree- or reset-based methods search over branches~\citep{ji2025treegrpo,samanta2026resets}, hierarchical and trajectory-level RL frameworks restructure the update itself~\citep{zhou2024archer,wang2025ragen}, and Fission-GRPO~\citep{zhang2026fission} synthesizes new recovery rollouts from failed trajectories. Denser step rewards instead come from process or hindsight evaluators~\citep{zhang2026webarbiter,tan2026hcapo}, turn-level shaping~\citep{xie2026tips,zeng2025turnlevel,gtpo2025,parthasarathi2025grpolambda}, or token-level reshaping~\citep{li2026oar,mesnard2021counterfactual,lai2024stepdpo}. A separate line places the user inside the RL loop: MUA-RL~\citep{zhao2025muarl} first trains agents against a simulated user interacting with live tools, and later methods add finer learning signals on top, such as UserRL~\citep{qian2025userrl} shaping turn- and trajectory-level rewards, SWEET-RL~\citep{zhou2025sweetrl} training a step-level critic from training-time information, and InfoPO~\citep{kong2026infopo} adding an information-gain reward from masked-context counterfactuals. All of these add signals beyond the sampled rollout: shaped rewards, a trained critic, an auxiliary intrinsic objective, or synthesized recovery rollouts. IAPO instead uses the dependencies already realized among the agent's own actions. It reweights the tokens of a rollout the baseline already sampled, using only the influence links realized inside that rollout.

\section{Problem Setup}
\label{sec:problem-setup}

We consider a multi-turn service episode in which user messages and tool returns progressively expose task state. Policy $\pi_\theta$ controls assistant replies and tool calls; user messages, tool returns, and rule feedback are observations used for provenance but receive no policy-gradient credit.

A trajectory alternates between observations and assistant actions:
\begin{equation}
  \tau = (o_1, a_1, o_2, a_2, \ldots, o_T, a_T),
  \label{eq:trajectory}
\end{equation}
where $o_t$ is an observation, $a_t$ is an assistant action, $L_i$ is the token length of action $i$, and $L=\sum_i L_i$. The environment returns terminal reward $R_\tau$. GRPO samples $\mathcal{G}=\{\tau_k\}_{k=1}^K$ for the same task and computes
\begin{equation}
  \hat A_\tau \;=\; \frac{R_\tau - \mu_{\mathcal{G}}}{\sigma_{\mathcal{G}} + \varepsilon},
  \label{eq:grpo-advantage}
\end{equation}
then broadcasts $\hat A_\tau$ to every action token.

\paragraph{Evolving-state credit assignment.} IAPO redistributes this scalar using only dependency evidence in the completed rollout: $A_{i,\ell}=\hat A_\tau w_i$, with $w_i>0$, $\sum_iL_iw_i=L$, and $w_i=1$ for an uninformative graph. The annotator extracts dependency labels but neither scores success nor creates an additional reward.

\section{Method}
\label{sec:method}

IAPO scores rollouts as GRPO does, extracts a signed influence graph from each completed rollout, and routes the trajectory advantage to each token as $A_{i,\ell}=\hat A_\tau\,w_i$ (Figure~\ref{fig:principle}). The routing weight $\mathbf w=(w_1,\ldots,w_T)$ redistributes the advantage across actions according to their influence: toward information that is used downstream, and toward the observable errors behind a negative outcome. $\mathbf w$ is
positive, bounded, and length-normalized, so routing reshapes how the advantage is shared without changing its total or its sign. The rest of this section builds $\mathbf w$ to this shape.

\subsection{Influence-Dependency Graph}
\label{sec:dependency-graph}

The map $\mathbf w$ is computed from observable evidence in the completed rollout. For $\tau$, let $V_\tau=\{a_1,\ldots,a_T\}$ contain the trainable assistant actions. IAPO builds the influence-dependency graph $G_\tau=(V_\tau,E_\tau^{+},E_\tau^{-})$, where $E_\tau^{+},E_\tau^{-}\subseteq\{(i,j):1\le i<j\le T\}$ and the sign of an edge is independent of the sign of $\hat A_\tau$. We use $+$ for support-use quantities and $-$ for error-evidence quantities throughout this section ($\boldsymbol \phi^{\boldsymbol \pm}$, $\mathbf {m^{\boldsymbol \pm}}$, and $\mathbf {w^{\boldsymbol\pm}}$ below). An edge $i\!\to^{+}\!j$ is a \emph{support-use edge}: $a_j$ consumes information supplied or elicited by $a_i$ and the annotation type is \texttt{input} or \texttt{success\_utilization}. An edge $i\!\to^{-}\!j$ is a \emph{failed-use edge}: $a_j$ repeats an observable error from $a_i$, uses its invalid output, or must fall back because of it, with annotation type \texttt{failed\_utilization}. The error indicator $\errflag_i\in\{0,1\}$ marks whether $a_i$ is error-bearing: a tool, schema, runtime, or rule failure, or an explicit user or environment rejection. A goal revision is not an error unless it rejects the preceding action, so $d_i^{-}>0$ implies $\errflag_i=1$.

An edge $i\!\to\!j$ is accepted when three conditions are jointly met:
\begin{enumerate}
  \item \textbf{Necessity.} $a_j$ references a specific information unit (an entity value, a tool-returned field, or an error trace) that originates from $a_i$ or from an observation elicited by $a_i$. Sharing the same topic or domain is insufficient: removing the referenced information unit must leave $a_j$'s argument or decision unsupported.
  \item \textbf{Explicit binding.} The referenced information unit is bound to a concrete element in $a_j$: an intent parameter, a tool-call argument, a conditional branch, or an error-recovery action. Abstract thematic similarity does not qualify.
  \item \textbf{Source priority.} When multiple prior actions supply overlapping information, the most comprehensive source is kept; ties are broken by earliest occurrence. Multiple sources are cited only when they contribute non-overlapping information units.
\end{enumerate}

User turns and tool returns provide evidence but never receive advantage. Their source type is retained for provenance resolution. A user reply is resolved to the assistant action that elicited it, while a tool return is resolved to the tool call that produced it. If a cited user turn has no assistant elicitor, it contributes only a direct-input count for the consuming action, written $u_i^{\mathrm{dir}}$. Duplicate user/elicitor evidence is counted once. Writing $d_i^{+}=|\{j: i \xrightarrow{+} j\}|$ and $d_i^{-}=|\{j: i \xrightarrow{-} j\}|$ for the number of outgoing support-use and failed-use edges from $a_i$, and $d_i^{\mathrm{in}}=u_i^{\mathrm{dir}} + |\{k: k \xrightarrow{+} i\}|$ for the support evidence $a_i$ consumes, whether from a direct user turn or an incoming support-use edge, these counts feed the graph features
\begin{equation}
  \phi_i^{+} = \log(1+d_i^{+}+d_i^{\mathrm{in}}),\quad
  \phi_i^{-} = \log\!\bigl(1+\errflag_i(1+d_i^{-}+d_i^{\mathrm{in}})\bigr).
  \label{eq:phi}
\end{equation}
The support score $\phi^{+}_i$ grows with the number of downstream consumers and upstream sources of $a_i$. The error score $\phi^{-}_i$ is zero unless $\errflag_i{=}1$; when an observable error is present, the inner unit term counts the error itself, $d_i^{-}$ counts its downstream failed use, and $d_i^{\mathrm{in}}$ counts the support evidence consumed by the erroneous action.

\subsection{From Features to Bounded Weights}
\label{sec:axioms}

Additive token corrections can change the sign or the total of the advantage, so IAPO uses multiplicative weights and turns each feature into a bounded weight around one. For a feature $\boldsymbol \phi\in\mathbb{R}_{\ge0}^{T}$, we standardize it by its length-weighted mean and deviation, clip at $c$, and renormalize so the length-weighted mean is one:
\begin{equation}
m_i=\frac{\exp\!\bigl(\mathrm{clip}(z_i,-c,c)\bigr)}{\tfrac{1}{L}\sum_j L_j\exp\!\bigl(\mathrm{clip}(z_j,-c,c)\bigr)},\qquad
z_i=\frac{\phi_i-\bar\phi}{\sigma_\phi+\varepsilon},
\label{eq:norm}
\end{equation}
where $\bar\phi=\tfrac{1}{L}\sum_i L_i\phi_i$, $\sigma_\phi^{2}=\tfrac{1}{L}\sum_i L_i(\phi_i-\bar\phi)^{2}$, and $\varepsilon=10^{-6}$. A feature above its mean gets weight above one and below its mean below one, and $c$ caps how far a step can move so no single hub dominates the update. Applying this to the graph features gives the support weight $m_i^{+}$ from $\boldsymbol\phi^{+}$ and the error weight $m_i^{-}$ from $\boldsymbol\phi^{-}$. Every token in action $i$ receives the same $w_i$, so the action carries token mass $L_iw_i$. The length-aware properties are proved in Appendix~B.

\subsection{Sign-Conditioned Routing}
\label{sec:routing}

\paragraph{Positive-advantage rollouts ($\hat A_\tau\!>\!0$).}
A positive-advantage rollout can still contain steps that made errors later steps had to repair, and IAPO gives them less credit than clean steps with the same support role. Starting from the support weight $m_i^{+}$, it discounts a step by its above-average error evidence $\psi_i^{-}=\phi_i^{-}-\bar\phi^{-}$, with $\bar\phi^{-}=\tfrac{1}{L}\sum_i L_i\phi_i^{-}$, and renormalizes:
\begin{equation}
w_i^{+}(\beta_{+}) = \frac{L\, m_i^{+}\, \exp(-\beta_{+}\psi_i^{-})}{\sum_j L_j\, m_j^{+}\, \exp(-\beta_{+}\psi_j^{-})}.
\label{eq:w-pos}
\end{equation}
This is the closed-form solution of the length-mean-constrained routing problem derived in Appendix~B. At $\beta_{+}{=}0$ it is $w_i^{+}{=}m_i^{+}$ (support only); a larger $\beta_{+}$ moves credit off repaired-error steps, and the denominator restores the length-weighted mean one.

\paragraph{Negative-advantage rollouts ($\hat A_\tau\!<\!0$).}
A negative-advantage rollout concentrates its advantage on the steps that produced the errors: with many errors and hard attribution, support reuse need not be at fault, so IAPO conservatively routes blame to error evidence only, riding on the error feature directly:
\begin{equation}
w_i^{-}=m_i^{-}.
\label{eq:w-neg}
\end{equation}
The error score $\phi^{-}$ is nonzero only after an observable error, and the normalizer keeps every multiplier positive, so more of the negative advantage reaches the steps with stronger error evidence. A projected support-prior alternative is evaluated in ablation and defined in Appendix~B.

\paragraph{Routed advantage.}
The final weight follows the trajectory's sign, $\boldsymbol w=\boldsymbol w^{\boldsymbol +}$ when $\hat A_\tau>0$ and $\boldsymbol w=\boldsymbol w^{\boldsymbol -}$ when $\hat A_\tau<0$, so the token advantage
\begin{equation}
A_{i,\ell}=\hat A_\tau\,w_i
\label{eq:iapo}
\end{equation}
is zero when $\hat A_\tau{=}0$ and is passed to the same clipped-loss implementation, replacing the flat per-token advantage used by GRPO. $\beta_{+}$ and $c$ are set in the experimental setup, and the projected negative branch is an ablation. After each rollout, a frozen annotator extracts the dependency graph, cached for the corresponding policy update.

\begin{table*}[t!]
  \centering
  \small
  \setlength{\tabcolsep}{5.0pt}
  \setlength{\aboverulesep}{0.1ex}
  \setlength{\belowrulesep}{0.1ex}
  \begin{tabular}{llcrrrrr}
    \toprule
    Model & Method & Think & $\tau^2$ (\%) & UserBench (\%) & AgentChangeBench (\%) & BFCL-MT (\%) & Avg (\%) \\
    \midrule
    GPT-4o-mini & -- & -- & 23.75 & 28.30 & 35.50 & 37.00 & 31.14 \\
    Qwen3.5-397B & Base & w/ & 84.38 & 73.92 & 43.50 & 64.00 & 66.45 \\
    \midrule
    \multirow{6}{*}{Qwen3-4B}
      & Base & w/o & 19.17 & 22.04 & 17.50 & 31.38 & 22.52 \\
      & Base & w/ & 31.46 & 25.33 & 25.00 & 30.50 & 28.07 \\
      & GRPO & w/o & \ms{34.79}{1.05} & \ms{23.27}{0.47} & \ms{21.50}{1.81} & \ms{31.69}{0.13} & 27.81 \\
      & GiGPO & w/o & \ms{37.22}{1.04} & \ms{25.06}{0.25} & \ms{24.42}{1.13} & \ms{\textbf{32.42}}{0.56} & 29.78 \\
      & InfoPO & w/o & \ms{37.99}{1.75} & \ms{20.47}{1.10} & \ms{17.56}{1.32} & \ms{29.94}{0.34} & 26.49 \\
      \cmidrule(l){2-8}
      & \textbf{IAPO (Ours)} & w/o & \ms{\textbf{38.40}}{1.63} & \ms{\textbf{26.89}}{0.61} & \ms{\textbf{28.06}}{2.58} & \ms{31.89}{0.97} & \textbf{31.31} \\
    \midrule
    \multirow{7}{*}{Qwen3-8B}
      & Base & w/o & 24.17 & 18.04 & 27.50 & 39.38 & 27.27 \\
      & Base & w/ & 37.29 & 22.08 & 29.00 & 39.00 & 31.84 \\
      & MUA-RL$^\dagger$ & w/o & 35.63 & 17.31 & 30.05 & 24.75 & 26.94 \\
      & GRPO & w/o & \ms{29.61}{3.56} & \ms{19.83}{0.19} & \ms{26.03}{1.79} & \ms{39.04}{0.71} & 28.63 \\
      & GiGPO & w/o & \ms{35.83}{1.04} & \ms{22.77}{0.21} & \ms{29.25}{0.88} & \ms{\textbf{40.00}}{0.44} & 31.96 \\
      & InfoPO & w/o & \ms{33.29}{0.77} & \ms{23.37}{0.38} & \ms{28.61}{1.44} & \ms{38.01}{0.63} & 30.82 \\
      \cmidrule(l){2-8}
      & \textbf{IAPO (Ours)} & w/o & \ms{\textbf{42.18}}{2.55} & \ms{\textbf{24.44}}{0.15} & \ms{\textbf{30.44}}{2.20} & \ms{39.36}{0.95} & \textbf{34.11} \\
    \midrule
    \multicolumn{8}{l}{\emph{External reported numbers$^{\ddagger}$ (quoted from prior work trained on larger, synthetically constructed corpora)}} \\
    ToolACE-2-8B    & SFT & -- & 20.53 & -- & -- & 37.00 & -- \\
    BitAgent-8B     & SFT & -- & 16.70 & -- & -- & 37.75 & -- \\
    Fission-GRPO-4B & RL  & w/ & 34.50 & -- & -- & 40.87 & -- \\
    Fission-GRPO-8B & RL  & w/ & 41.27 & -- & -- & 46.75 & -- \\
    \bottomrule
  \end{tabular}
  \caption{Main results (\%). $\pm$ is the standard deviation over three seeds. $^\dagger$: results re-evaluated using publicly available checkpoint. $^{\ddagger}$: results copied from the published papers. IAPO uses only the \textbf{178-task }$\tau^2$-Bench training split, while Fission-GRPO~\citep{zhang2026fission} uses \textbf{630 tasks across 11 domains grounded in  BFCL characteristics}, designed and cleaned with Claude Sonnet 4. ToolACE-2-8B~\citep{liu2025toolace} and BitAgent-8B~\citep{bitagent2025} are specialized function-calling agents.
}
  \label{tab:main}
\end{table*}

\subsection{Properties}
\label{sec:properties}

The routing layer has four safeguards; proofs are in Appendix~B.
\begin{proposition}[Mass conservation]
$\sum_i\sum_{\ell\in a_i} A_{i,\ell}=L\hat A_\tau$. IAPO redistributes the existing rollout advantage without changing its total token mass.
\end{proposition}

\begin{proposition}[Uniform fallback]
If the active graph features are constant, then $m_i^{\pm}=1$, $w_i=1$, and Eq.~\eqref{eq:iapo} assigns the same advantage to every token. With no observable error, $\phi^{-}_i$ is constant and the negative branch remains uniform.
\end{proposition}

\begin{proposition}[Sign preservation]
Because all routing multipliers are positive, every token advantage has the same sign as $\hat A_\tau$.
\end{proposition}

\begin{proposition}[Optimization-interface preservation]
IAPO preserves the terminal reward, group normalization, prompts, tools, rollout sampling, and optimization interface while intentionally changing the token-weighted policy-gradient estimator.
\end{proposition}

\section{Experiments}
\label{sec:experiments}

We evaluate IAPO through a controlled comparison with GRPO~\citep{shao2024deepseekmath} and complementary comparisons with GiGPO~\citep{feng2026gigpo} and InfoPO~\citep{kong2026infopo}. GRPO~\citep{shao2024deepseekmath} and IAPO share the same training protocol; their only difference is the trajectory-to-token advantage map. The evaluation covers success rate, per-domain breakdown, out-of-domain transfer, and annotator stability.

\subsection{Experimental Setup}
\label{sec:exp-setup}

\textbf{Benchmarks.} We train on the $\tau^2$-Bench~\citep{barres2025tau2bench} training split (airline/retail/telecom, 178 tasks) and evaluate on:
\begin{itemize}
  \item $\tau^2$-Bench held-out test (20/40/40 tasks $\times$ 4 trials; domain-macro pass$^1$);
  \item UserBench~\citep{qian2025userbench} travel22/33/44 (101/87/67 cases; official multi-choice Score);
  \item AgentChangeBench~\citep{rana2025agentchangebench} banking/education (100/100 tasks; TSR);
  \item BFCL-v4 Multi-Turn (function-calling retention)~\citep{patil2023gorilla}.
\end{itemize}

\textbf{Baselines.} GRPO~\citep{shao2024deepseekmath} is the controlled baseline, differing from IAPO only in the trajectory-to-token advantage map. GiGPO~\citep{feng2026gigpo} adds cross-rollout step credit by grouping actions with similar preceding observations; we use the authors' official implementation with similarity-based grouping ($\gamma{=}0.95$, threshold $0.9$, and $\omega{=}1.0$), without further tuning. InfoPO~\citep{kong2026infopo} estimates per-turn information acquisition by comparing policy likelihoods under observed and masked contexts, fusing the resulting intrinsic reward with outcome advantage. The released MUA-RL-8B~\citep{zhao2025muarl} is included as an external checkpoint comparator; it was trained on $\tau^1$ retail/airline with a GPT-4o user simulator and is evaluated without further optimization. For broader context, we also report Fission-GRPO~\citep{zhang2026fission}, ToolACE-2-8B~\citep{liu2025toolace}, and BitAgent-8B~\citep{bitagent2025}. Fission-GRPO~\citep{zhang2026fission} adds an error-recovery loop to GRPO: an error simulator diagnoses failed on-policy trajectories, constructs corrective samples, and triggers additional rollout batches for corrective training. ToolACE-2-8B~\citep{liu2025toolace} and BitAgent-8B~\citep{bitagent2025} are fine-tuned tool-use agents.

\textbf{Training and evaluation.} We train Qwen3-4B and Qwen3-8B with thinking disabled on $8{\times}$ NVIDIA H20 GPUs using a verl/FSDP GRPO stack~\citep{sheng2024verl,zhao2023pytorch}, $N{=}16$ samples/prompt, max 30 turns. Dependency annotation uses Qwen3-32B at temperature~0. Interactive evaluation follows the $\tau^2$-Bench protocol with held-out GPT-5.2 user simulators~\citep{barres2025tau2bench}. IAPO defaults: $\beta_{+}{=}0.05$, $w^{-}{=}m^{-}$, $c{=}0.25$. For each seed, we report the mean over the final three checkpoints; full hyperparameters and compute budget are in Appendix~F.

\begin{table}[t]
  \centering
  \small
  \setlength{\tabcolsep}{4.5pt}
  \begin{tabular}{@{}llrrr@{}}
    \toprule
    Benchmark & Split & GRPO & IAPO & $\Delta$ \\
    \midrule
    \multirow{3}{*}{$\tau^2$}
      & Airline & $22.78$ & $29.72$ & $+6.94$ \\
      & Retail  & $44.31$ & $56.88$ & $+12.57$ \\
      & Telecom & $21.74$ & $39.93$ & $\mathbf{+18.19}$ \\
    \midrule
    \multirow{3}{*}{UserBench}
      & T22 & $20.68$ & $25.65$ & $+4.97$ \\
      & T33 & $19.41$ & $24.50$ & $\mathbf{+5.09}$ \\
      & T44 & $19.10$ & $22.55$ & $+3.45$ \\
    \midrule
    \multirow{2}{*}{AgentChange}
      & Banking   & $28.23$ & $32.67$ & $\mathbf{+4.44}$ \\
      & Education  & $23.82$ & $28.22$ & $+4.40$ \\
    \bottomrule
  \end{tabular}
  \caption{Per-split results at 8B (three-seed mean). }
  \label{tab:domain-results}
\end{table}

\subsection{Main Results}
\label{sec:main-results}

Table~\ref{tab:main} shows IAPO's largest improvement on $\tau^2$-Bench: at 8B, it raises domain-macro pass$^1$ from $29.61\%$ to $42.18\%$, a $+12.57$ pp gain over matched GRPO. The same update improves UserBench by $+4.61$ pp and AgentChangeBench by $+4.41$ pp, while BFCL-MT remains comparable ($39.36\%$ vs.\ $39.04\%$). Crucially, these in-domain gains do not come at the cost of out-of-distribution behavior: BFCL-MT stays effectively flat relative to GRPO at both scales, indicating that IAPO's service-agent advantage reflects sharper credit assignment rather than overfitting away function-calling ability (Table~\ref{tab:main}).

\textbf{Delayed state use yields the largest gain.} Table~\ref{tab:domain-results} resolves the aggregate results by split. On $\tau^2$-Bench, IAPO improves airline from $22.78$ to $29.72$ ($+6.94$ pp), retail from $44.31$ to $56.88$ ($+12.57$ pp), and telecom from $21.74$ to $39.93$ ($+18.19$ pp). Telecom requires account or line state to persist until a later clarification determines its use; this is the setting in which a flat rollout advantage provides the least specific feedback about earlier actions.

\textbf{Generalization.} IAPO is trained only on $\tau^2$-Bench, yet the same table shows it improving every out-of-domain split. UserBench requires accumulating preferences that arrive turn by turn: IAPO gains $+4.97$, $+5.09$, and $+3.45$ pp across the three travel splits. AgentChangeBench changes the desired outcome mid-interaction: IAPO gains $+4.44$ pp on banking and $+4.40$ pp on education, remaining ahead where account state must survive the goal shift. The higher success also comes with less redundant tool use: at 4B, IAPO lowers macro tool-call redundancy from GRPO's $46.00$ to $29.74$ (Appendix~F), so the gain reflects reuse of acquired state instead of repeated queries.

\textbf{The router is far from flat.} The improvement comes from a routing map that departs substantially from uniform credit. Across the audited action steps, the positive weight $w^{+}$ has standard deviation $0.222$ and spans $0.637$--$1.505$, and $86.5\%$ of steps move by more than $10\%$ from flat credit (Appendix~E). The negative branch is deliberately sparser, with $27.1\%$ of steps exceeding the same threshold, consistent with $\phi^{-}$ activating only after an observable error. IAPO changes the per-action advantage while departing substantially from the flat GRPO update.

\subsection{Ablation and Sensitivity}
\label{sec:ablation}

\begin{table}[t]
\centering\small\setlength{\tabcolsep}{2pt}
\begin{tabular}{@{}llccc@{}}
\toprule
Axis & Setting & $\tau^2$ & UB & AC \\
\midrule
Ref. & GRPO & \ms{29.61}{3.56} & \ms{19.83}{0.19} & \ms{26.03}{1.79} \\
\midrule
$\beta_{+}$ & $0$ & 32.50 & 20.92 & 27.58 \\
 & \textbf{$.05$ def.} & \ms{\textbf{42.18}}{2.55} & \ms{24.44}{0.15} & \ms{30.44}{2.20} \\
 & $.1$ & 33.96 & \textbf{25.33} & \textbf{30.92} \\
 & $.3$ & 31.39 & 24.08 & 28.83 \\
\midrule
$c$ & $.125$ & 32.92 & 19.11 & 30.17 \\
 & \textbf{$.25$ def.} & \ms{42.18}{2.55} & \ms{\textbf{24.44}}{0.15} & \ms{\textbf{30.44}}{2.20} \\
 & $.5$ & \textbf{42.29} & 17.78 & 25.67 \\
\midrule
$w^{-}$ & \textbf{cons. def.} & \ms{\textbf{42.18}}{2.55} & \ms{\textbf{24.44}}{0.15} & \ms{\textbf{30.44}}{2.20} \\
 & proj. $0$ & 41.88 & 23.87 & 28.33 \\
 & proj. $.05$ & 41.46 & 23.16 & 28.58 \\
\midrule
Annot. & \textbf{Q3-32B def.} & \ms{\textbf{42.18}}{2.55} & \ms{\textbf{24.44}}{0.15} & \ms{30.44}{2.20} \\
 & Gemma4-31B & 40.90 & 24.45 & \textbf{33.25} \\
\bottomrule
\end{tabular}
  \caption{Ablation of routing hyperparameters and the frozen annotator on Qwen3-8B (\%).}
  \label{tab:ablation-summary}
\end{table}

Each panel in Table~\ref{tab:ablation-summary} isolates one routing decision, either by disabling it or by sweeping its value: $\beta_{+}$ controls how repaired errors affect positive-advantage rollouts, $c$ bounds the weight spread, $w^{-}$ determines whether failure credit begins from support or observed error, and the annotator swap tests transfer across model families.

\textbf{Support and error each carry useful signal.} With $\beta_{+}{=}0$, positive-advantage rollouts route credit by support use and negative-advantage ones route it by error propagation. This stripped rule already raises $\tau^2$ from $29.61$ to $32.50$. A small positive-branch gate ($\beta_{+}{=}0.05$) reaches $42.18$ by reducing positive credit on actions whose earlier errors were repaired downstream. Within the $\beta_{+}$ sweep, larger gates yield $33.96$ and $31.39$ on $\tau^2$, indicating that error evidence refines support routing without taking its place.

\textbf{Clipping prevents hubs from dominating the update.} At $c{=}0.125$, routing stays near flat credit and reaches $32.92$ on $\tau^2$. Increasing the bound to $c{=}0.5$ raises the in-domain score to $42.29$, but UserBench and AgentChangeBench fall to $17.78$ and $25.67$. The default $c{=}0.25$ gives the strongest joint in-domain and transfer profile.

\textbf{Failure routing follows error propagation, not support centrality.} The conservative branch begins from observable error and reaches $42.18$ on $\tau^2$ and $30.44$ on AgentChangeBench. Both projected variants start from the support prior and are lower on these measures. The two panels rule out degree-based explanations: positive and negative advantage follow different evidence.

\textbf{The gain transfers across annotator families.} Replacing Qwen3-32B with Gemma4-31B yields $40.90$ on $\tau^2$, $24.45$ on UserBench, and $33.25$ on AgentChangeBench. This transfer and the routing correlations in Table~\ref{tab:annotator} indicate that IAPO does not rely on the default Qwen annotator's particular outputs.

\begin{table}[t]
\centering\scriptsize
\setlength{\tabcolsep}{1.5pt}
\begin{tabular*}{\columnwidth}{@{\extracolsep{\fill}}lcccccc@{}}
\toprule
\multicolumn{1}{c}{Annotator} & \multicolumn{3}{c}{Edge agreement} & \multicolumn{3}{c}{Routing agreement}\\
\cmidrule(lr){2-4}\cmidrule(l){5-7}
 & F1$_e$ & Jaccard & Cohen's $\kappa$ & $\rho(m^+)$ & $\rho(m^-)$ & Sign agreement \\
\midrule
Qwen3-32B (ref.)  & 0.852 & 0.822 & 0.812 & -- & -- & -- \\
\midrule
Qwen3.5-27B       & 0.875 & 0.846 & 0.833 & \textbf{0.769} & \textbf{1.000} & 0.858 \\
Gemma4-31B        & 0.878 & \textbf{0.857} & \textbf{0.849} & 0.745 & 0.996 & 0.858 \\
\midrule
DeepSeek-V4-Flash & \textbf{0.879} & 0.852 & 0.843 & 0.758 & 0.975 & \textbf{0.877} \\
Qwen3-235B-A22B   & 0.860 & 0.836 & 0.809 & 0.732 & 0.984 & 0.848 \\
Qwen3.5-397B-A17B & 0.852 & 0.818 & 0.805 & 0.726 & 0.962 & 0.834 \\
\midrule
Mean              & 0.869 & 0.842 & 0.828 & 0.746 & 0.983 & 0.855 \\
\bottomrule
\end{tabular*}
\caption{Cross-model annotation consistency. Edge agreement reports F1, Jaccard, and Cohen's $\kappa$; routing agreement reports Spearman correlations with the Qwen3-32B reference and sign agreement of the routed advantages.}
\label{tab:annotator}
\end{table}

\subsection{Annotator Stability and Behavioral Analysis}
\label{sec:stability}

\textbf{Cross-annotator stability.} We audit all $178$ $\tau^2$-Bench training tasks by rolling out the untrained Qwen3-8B policy once per task, yielding $2{,}349$ annotated action nodes across airline, retail, and telecom. Six frozen annotators achieve $100\%$ parse rate. Their canonical edge sets show substantial set-level agreement (mean edge-typed F1 $0.87$, source Jaccard $0.84$, mean pairwise Cohen's $\kappa{=}0.83$). The routing quantities used by IAPO are also stable: against the Qwen3-32B reference, $m^{+}$ has Spearman $\rho=0.73$--$0.77$, while $m^{-}$ has $\rho=0.96$--$1.00$ (Table~\ref{tab:annotator}). We further audit the lowest-agreement nodes in Appendix E. Disagreements fall mainly into benign, interpretable sources: interchangeable boilerplate turns (greetings, closings) where annotators pick different but near-identical preceding turns, and attribution breadth on already-failed steps (e.g., hallucinated tool calls) where annotators differ in attribution breadth, while generally agreeing that a fault exists. Neither source changes the routing multipliers, so agreement holds at both the edge and induced-routing levels.

\begin{figure}[!t]
\centering
\includegraphics[width=0.60\columnwidth]{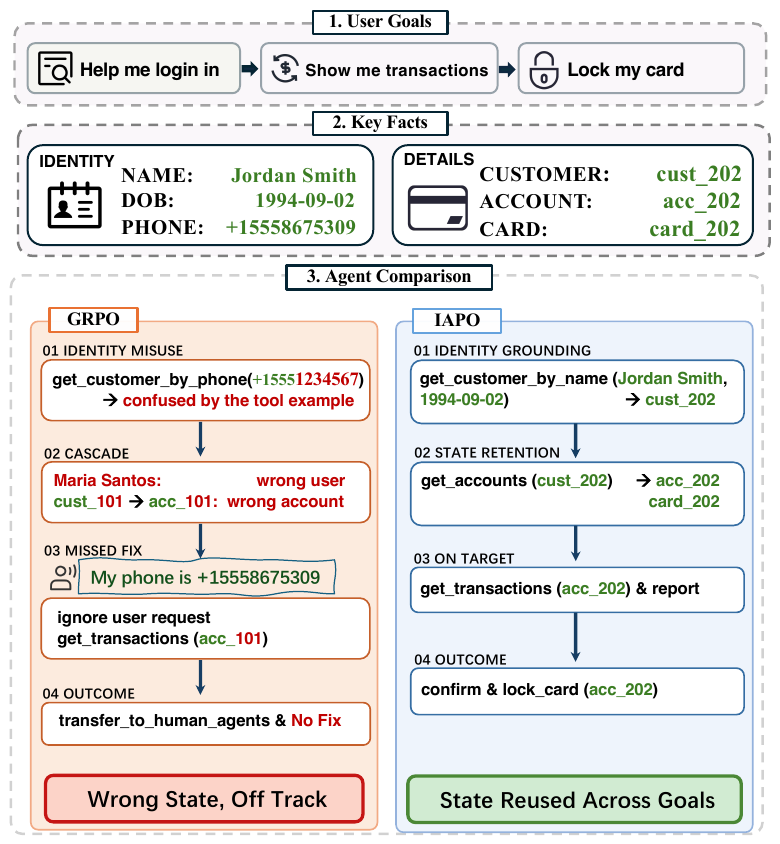}
\caption{Banking case study with three successive goal shifts. (Left) GRPO anchors on a schema-example identity and propagates the wrong account state. (Right) IAPO grounds the correct customer from supplied credentials and retains the linked state across goals.}
\label{fig:banking-goal-shift}
\end{figure}

\textbf{Behavioral case study.} Figure~\ref{fig:banking-goal-shift} illustrates a banking episode in AgentChangeBench~\citep{rana2025agentchangebench} whose goals shift from authentication to transaction review and fraud response. Before receiving identity information, GRPO uses the example phone exposed in the tool schema, obtains Maria Santos ($\texttt{cust\_101}$), and propagates the resulting account state ($\texttt{acc\_101}$) into its first transaction query. When the user later supplies the actual phone number, GRPO does not re-ground the customer and instead transfers to a human. IAPO first grounds Jordan Smith from the supplied name and date of birth, then retains the linked state ($\texttt{cust\_202}$, $\texttt{acc\_202}$, and $\texttt{card\_202}$) across goal shifts, eventually locking $\texttt{card\_202}$. The case study isolates state grounding and cross-goal reuse.

\textbf{Function-calling retention evaluation.} We use BFCL-v4 Multi-Turn~\citep{patil2023gorilla} as a function-calling retention check. Table~\ref{tab:bfcl} reports BFCL sub-splits at both scales. Relative to the untrained base, IAPO's BFCL-MT score is essentially unchanged at 8B ($39.36$ vs.\ $39.38$) and higher at 4B ($31.89$ vs.\ $31.38$), so interactive training leaves out-of-distribution function calling intact.

\begin{table}[!htb]
\centering\small
\setlength{\tabcolsep}{1.5pt}
\begin{tabular*}{\columnwidth}{@{\extracolsep{\fill}}llrrrrr@{}}
\toprule
Model & Method & \shortstack{Multi-\\Turn} & Base & \shortstack{Multi-\\Function} & \shortstack{Multi-\\Parameter} & Long \\
\midrule
4B & Base (w/)   & 31.38 & \textbf{43.50} & 27.00 & 28.50 & 26.50 \\
   & GRPO   & 31.69 & 39.61 & 30.67 & 28.28 & 28.22 \\
   & GiGPO  & \textbf{32.42} & 31.83 & \textbf{32.33} & \textbf{32.67} & \textbf{32.83} \\
   & InfoPO & 29.94 & 37.94 & 29.00 & 26.50 & 26.33 \\
   & IAPO   & 31.89 & 40.06 & 30.78 & 28.28 & 28.44 \\
\midrule
8B & Base (w/)  & 39.38 & 49.00 & 37.50 & \textbf{41.50} & 29.50 \\
   & MUA-RL$^\dagger$ & 24.75 & 35.50 & 21.50 & 22.00 & 20.00 \\
   & GRPO   & 39.04 & 48.94 & \textbf{40.67} & 34.78 & 31.78 \\
   & GiGPO  & \textbf{40.00} & 40.17 & 39.67 & 40.33 & \textbf{39.83} \\
   & InfoPO & 38.01 & \textbf{49.39} & 37.44 & 30.78 & 34.44 \\
   & IAPO   & 39.36 & 48.33 & 39.61 & 36.00 & 33.50 \\
\bottomrule
\end{tabular*}
\caption{BFCL-v4 Multi-Turn sub-split accuracy (\%).}
\label{tab:bfcl}
\end{table}

\section{Conclusion}
\label{sec:conclusion}

In this work, we proposed IAPO, a novel RL algorithm to tackle the credit assignment challenge in evolving-state multi-turn service-agent training. IAPO introduces a typed influence-dependency graph that enables fine-grained per-step credit assignment while retaining the original trajectory-level outcome signal. By extracting support-use and failed-use dependencies from completed rollouts, it achieves this without replacing the final outcome with an independently learned local reward or requiring counterfactual continuations. Across three interactive service-agent benchmarks, IAPO consistently improves over GRPO and competitive credit-assignment baselines. The gains are largest in interactions with rich user and tool exchanges, consistent with the intuition that dense dependencies among actions are where uniform trajectory-level credit is least informative. These results establish within-rollout dependency structure as a practical source of selective credit for sparse-outcome multi-turn agent training.



\begin{thebibliography}{55}
\providecommand{\natexlab}[1]{#1}

\bibitem[{Arjona-Medina et~al.(2019)Arjona-Medina, Gillhofer, Widrich,
  Unterthiner, Brandstetter, and Hochreiter}]{arjonamedina2019rudder}
Arjona-Medina, J.~A.; Gillhofer, M.; Widrich, M.; Unterthiner, T.;
  Brandstetter, J.; and Hochreiter, S. 2019.
\newblock {RUDDER}: Return Decomposition for Delayed Rewards.
\newblock In \emph{Advances in Neural Information Processing Systems},
  volume~32.

\bibitem[{Barres et~al.(2025)Barres, Dong, Ray, Si, and
  Narasimhan}]{barres2025tau2bench}
Barres, V.; Dong, H.; Ray, S.; Si, X.; and Narasimhan, K. 2025.
\newblock $\tau^2$-Bench: Evaluating Conversational Agents in a Dual-Control
  Environment.
\newblock arXiv:2506.07982.

\bibitem[{{BitAgent}(2025)}]{bitagent2025}
{BitAgent}. 2025.
\newblock {BitAgent-8B}.
\newblock \url{https://huggingface.co/BitAgent/BitAgent-8B}.
\newblock Commit hash ca31a77.

\bibitem[{Budzianowski et~al.(2018)Budzianowski, Wen, Tseng, Casanueva, Ultes,
  Ramadan, and Gasic}]{budzianowski2018multiwoz}
Budzianowski, P.; Wen, T.-H.; Tseng, B.-H.; Casanueva, I.; Ultes, S.; Ramadan,
  O.; and Gasic, M. 2018.
\newblock {MultiWOZ} -- A Large-Scale Multi-Domain Wizard-of-{O}z Dataset for
  Task-Oriented Dialogue Modelling.
\newblock In \emph{Proceedings of the 2018 conference on empirical methods in
  natural language processing}, 5016--5026.

\bibitem[{Cui et~al.(2025)Cui, Yuan, Wang, Wang, Zhang, Chen, Li, He, Fan, Yu
  et~al.}]{cui2025prime}
Cui, G.; Yuan, L.; Wang, Z.; Wang, H.; Zhang, Y.; Chen, J.; Li, W.; He, B.;
  Fan, Y.; Yu, T.; et~al. 2025.
\newblock Process reinforcement through implicit rewards.
\newblock arXiv:2502.01456.

\bibitem[{DeepSeek-AI et~al.(2025)DeepSeek-AI, Liu, Feng, Xue, Wang, Wu, Lu,
  Zhao, Deng, Zhang, Ruan et~al.}]{liu2024deepseekv3}
DeepSeek-AI; Liu, A.; Feng, B.; Xue, B.; Wang, B.; Wu, B.; Lu, C.; Zhao, C.;
  Deng, C.; Zhang, C.; Ruan, C.; et~al. 2025.
\newblock DeepSeek-V3 Technical Report.
\newblock arXiv:2412.19437.

\bibitem[{Deng et~al.(2023)Deng, Gu, Zheng, Chen, Stevens, Wang, Sun, and
  Su}]{deng2023mind2web}
Deng, X.; Gu, Y.; Zheng, B.; Chen, S.; Stevens, S.; Wang, B.; Sun, H.; and Su,
  Y. 2023.
\newblock Mind2web: Towards a generalist agent for the web.
\newblock In \emph{Advances in Neural Information Processing Systems},
  volume~36, 28091--28114.

\bibitem[{Ding et~al.(2026)Ding, Le, Han, Ruan, Jin, Kumar, Wang, and
  Deoras}]{gtpo2025}
Ding, Y.; Le, H.; Han, S.; Ruan, K.; Jin, Z.; Kumar, V.; Wang, Z.; and Deoras,
  A. 2026.
\newblock Empowering multi-turn tool-integrated agentic reasoning with group
  turn policy optimization.
\newblock In \emph{Proceedings of the 64th Annual Meeting of the Association
  for Computational Linguistics (Volume 1: Long Papers)}, 42409--42423.

\bibitem[{Feng et~al.(2026)Feng, Xue, Liu, and An}]{feng2026gigpo}
Feng, L.; Xue, Z.; Liu, T.; and An, B. 2026.
\newblock Group-in-group policy optimization for llm agent training.
\newblock \emph{Advances in Neural Information Processing Systems}, 38:
  46375--46408.

\bibitem[{Harutyunyan et~al.(2019)Harutyunyan, Dabney, Mesnard, Azar, Piot,
  Heess, van Hasselt, Wayne, Singh, Precup et~al.}]{harutyunyan2019hindsight}
Harutyunyan, A.; Dabney, W.; Mesnard, T.; Azar, M.~G.; Piot, B.; Heess, N.; van
  Hasselt, H.; Wayne, G.; Singh, S.; Precup, D.; et~al. 2019.
\newblock Hindsight Credit Assignment.
\newblock In \emph{Advances in Neural Information Processing Systems}.

\bibitem[{Ji et~al.(2025)Ji, Ma, Wang, Chen, Chu, and Wu}]{ji2025treegrpo}
Ji, Y.; Ma, Z.; Wang, Y.; Chen, G.; Chu, X.; and Wu, L. 2025.
\newblock Tree Search for LLM Agent Reinforcement Learning.
\newblock arXiv:2509.21240.

\bibitem[{Kong et~al.(2026)Kong, Zhang, Deng, Wu, Luo, and
  Liu}]{kong2026infopo}
Kong, F.; Zhang, J.; Deng, M.; Wu, C.; Luo, Y.; and Liu, B. 2026.
\newblock InfoPO: Information-Driven Policy Optimization for User-Centric
  Agents.
\newblock In \emph{Forty-third International Conference on Machine Learning}.

\bibitem[{Lai et~al.(2024)Lai, Tian, Chen, Yang, Peng, and
  Jia}]{lai2024stepdpo}
Lai, X.; Tian, Z.; Chen, Y.; Yang, S.; Peng, X.; and Jia, J. 2024.
\newblock Step-{DPO}: Step-Wise Preference Optimization for Long-Chain
  Reasoning of {LLM}s.
\newblock arXiv:2406.18629.

\bibitem[{Li et~al.(2023)Li, Zhao, Yu, Song, Li, Yu, Li, Huang, and
  Li}]{li2023apibank}
Li, M.; Zhao, Y.; Yu, B.; Song, F.; Li, H.; Yu, H.; Li, Z.; Huang, F.; and Li,
  Y. 2023.
\newblock {API-Bank}: A Comprehensive Benchmark for Tool-Augmented {LLM}s.
\newblock In \emph{Proceedings of the 2023 conference on empirical methods in
  natural language processing}, 3102--3116.

\bibitem[{Li et~al.(2026)Li, Kang, Xiao, Xing, Si, Li, Gong, Yang, Xiao, and
  Guo}]{li2026oar}
Li, Z.; Kang, L.; Xiao, F.; Xing, L.; Si, Q.; Li, Z.; Gong, W.; Yang, D.; Xiao,
  Y.; and Guo, H. 2026.
\newblock Outcome-Grounded Advantage Reshaping for Fine-Grained Credit
  Assignment in Mathematical Reasoning.
\newblock arXiv:2601.07408.

\bibitem[{Lightman et~al.(2024)Lightman, Kosaraju, Burda, Edwards, Baker, Lee,
  Leike, Schulman, Sutskever, and Cobbe}]{lightman2024verify}
Lightman, H.; Kosaraju, V.; Burda, Y.; Edwards, H.; Baker, B.; Lee, T.; Leike,
  J.; Schulman, J.; Sutskever, I.; and Cobbe, K. 2024.
\newblock Let's Verify Step by Step.
\newblock In \emph{International Conference on Learning Representations},
  volume 2024, 39578--39601.

\bibitem[{Liu et~al.(2025{\natexlab{a}})Liu, Huang, Zeng, Hao, Yu, Li, Wang,
  Gan, Liu, Yu, Wang, Wang, Ning, Hou, Wang, Wu, Wang, Liu, Wang, Tang, Tu,
  Shang, Jiang, Tang, Lian, Liu, and Chen}]{liu2025toolace}
Liu, W.; Huang, X.; Zeng, X.; Hao, X.; Yu, S.; Li, D.; Wang, S.; Gan, W.; Liu,
  Z.; Yu, Y.; Wang, Z.; Wang, Y.; Ning, W.; Hou, Y.; Wang, B.; Wu, C.; Wang,
  X.; Liu, Y.; Wang, Y.; Tang, D.; Tu, D.; Shang, L.; Jiang, X.; Tang, R.;
  Lian, D.; Liu, Q.; and Chen, E. 2025{\natexlab{a}}.
\newblock {ToolACE}: Winning the Points of {LLM} Function Calling.
\newblock In \emph{The Thirteenth International Conference on Learning
  Representations}.

\bibitem[{Liu et~al.(2024)Liu, Yu, Zhang, Xu, Lei, Lai, Gu, Ding, Men, Yang
  et~al.}]{liu2024agentbench}
Liu, X.; Yu, H.; Zhang, H.; Xu, Y.; Lei, X.; Lai, H.; Gu, Y.; Ding, H.; Men,
  K.; Yang, K.; et~al. 2024.
\newblock Agentbench: Evaluating llms as agents.
\newblock In \emph{International Conference on Learning Representations},
  volume 2024, 52989--53046.

\bibitem[{Liu et~al.(2025{\natexlab{b}})Liu, Chen, Li, Qi, Pang, Du, Lee, and
  Lin}]{liu2025drgrpo}
Liu, Z.; Chen, C.; Li, W.; Qi, P.; Pang, T.; Du, C.; Lee, W.~S.; and Lin, M.
  2025{\natexlab{b}}.
\newblock Understanding r1-zero-like training: A critical perspective.
\newblock arXiv:2503.20783.

\bibitem[{Mesnard et~al.(2021)Mesnard, Weber, Viola, Thakoor, Saade,
  Harutyunyan, Dabney, Stepleton, Heess, Guez
  et~al.}]{mesnard2021counterfactual}
Mesnard, T.; Weber, T.; Viola, F.; Thakoor, S.; Saade, A.; Harutyunyan, A.;
  Dabney, W.; Stepleton, T.; Heess, N.; Guez, A.; et~al. 2021.
\newblock Counterfactual Credit Assignment in Model-Free Reinforcement
  Learning.
\newblock In \emph{International Conference on Machine Learning}.

\bibitem[{Mialon et~al.(2024)Mialon, Fourrier, Wolf, LeCun, and
  Scialom}]{mialon2023gaia}
Mialon, G.; Fourrier, C.; Wolf, T.; LeCun, Y.; and Scialom, T. 2024.
\newblock {GAIA}: A Benchmark for General {AI} Assistants.
\newblock In \emph{International Conference on Learning Representations},
  volume 2024, 9025--9049.

\bibitem[{Ng, Harada, and Russell(1999)}]{ng1999rewardshaping}
Ng, A.~Y.; Harada, D.; and Russell, S. 1999.
\newblock Policy invariance under reward transformations: Theory and
  application to reward shaping.
\newblock In \emph{International Conference on Machine Learning}, volume~99,
  278--287. Citeseer.

\bibitem[{OpenAI et~al.(2024)OpenAI, Achiam, Adler, Agarwal, Ahmad, Akkaya,
  Aleman, Almeida, Altenschmidt, Altman, Anadkat et~al.}]{achiam2023gpt}
OpenAI; Achiam, J.; Adler, S.; Agarwal, S.; Ahmad, L.; Akkaya, I.; Aleman,
  F.~L.; Almeida, D.; Altenschmidt, J.; Altman, S.; Anadkat, S.; et~al. 2024.
\newblock GPT-4 Technical Report.
\newblock arXiv:2303.08774.

\bibitem[{Parthasarathi et~al.(2025)Parthasarathi, Reymond, Chen, Cui, and
  Chandar}]{parthasarathi2025grpolambda}
Parthasarathi, P.; Reymond, M.; Chen, B.; Cui, Y.; and Chandar, S. 2025.
\newblock {GRPO}-$\lambda$: Credit Assignment improves {LLM} Reasoning.
\newblock arXiv:2510.00194.

\bibitem[{Patil et~al.(2024)Patil, Zhang, Wang, and
  Gonzalez}]{patil2023gorilla}
Patil, S.~G.; Zhang, T.; Wang, X.; and Gonzalez, J.~E. 2024.
\newblock Gorilla: Large language model connected with massive apis.
\newblock In \emph{Advances in Neural Information Processing Systems},
  volume~37, 126544--126565.

\bibitem[{Qian et~al.(2025{\natexlab{a}})Qian, Liu, Prabhakar, Liu, Zhang,
  Chen, Ji, Yao, Heinecke, Savarese, Xiong, and Wang}]{qian2025userbench}
Qian, C.; Liu, Z.; Prabhakar, A.; Liu, Z.; Zhang, J.; Chen, H.; Ji, H.; Yao,
  W.; Heinecke, S.; Savarese, S.; Xiong, C.; and Wang, H. 2025{\natexlab{a}}.
\newblock {UserBench}: An Interactive Gym Environment for User-Centric Agents.
\newblock arXiv:2507.22034.

\bibitem[{Qian et~al.(2025{\natexlab{b}})Qian, Liu, Prabhakar, Qiu, Liu, Chen,
  Kokane, Ji, Yao, Heinecke et~al.}]{qian2025userrl}
Qian, C.; Liu, Z.; Prabhakar, A.; Qiu, J.; Liu, Z.; Chen, H.; Kokane, S.; Ji,
  H.; Yao, W.; Heinecke, S.; et~al. 2025{\natexlab{b}}.
\newblock {UserRL}: Training Interactive User-Centric Agent via Reinforcement
  Learning.
\newblock arXiv:2509.19736.

\bibitem[{Qin et~al.(2023)Qin, Liang, Ye, Zhu, Yan, Lu, Lin, Cong, Tang, Qian
  et~al.}]{qin2024toolllm}
Qin, Y.; Liang, S.; Ye, Y.; Zhu, K.; Yan, L.; Lu, Y.; Lin, Y.; Cong, X.; Tang,
  X.; Qian, B.; et~al. 2023.
\newblock {ToolLLM}: Facilitating Large Language Models to Master 16000+
  Real-World {API}s.
\newblock In \emph{The twelfth international conference on learning
  representations}.

\bibitem[{Rana et~al.(2025)Rana, Man, Msiiwa, Paine, Zhu, Dev, Sharma, and
  R}]{rana2025agentchangebench}
Rana, M.; Man, C.; Msiiwa, A.~E.; Paine, J.; Zhu, K.; Dev, S.; Sharma, V.; and
  R, A.~M. 2025.
\newblock {AgentChangeBench}: A Multi-Dimensional Evaluation Framework for
  Goal-Shift Robustness in Conversational AI.
\newblock arXiv:2510.18170.

\bibitem[{Samanta et~al.(2026)Samanta, Magesh, Jain, Yu, Jiang, Asadi, Hassani,
  Sajda, Bhandari, and Efroni}]{samanta2026resets}
Samanta, A.; Magesh, A.; Jain, A.; Yu, Y.; Jiang, D.; Asadi, K.; Hassani, K.;
  Sajda, P.; Bhandari, J.; and Efroni, Y. 2026.
\newblock Credit Assignment with Resets in Language Model Reasoning.
\newblock arXiv:2605.25507.

\bibitem[{Schick et~al.(2023)Schick, Dwivedi-Yu, Dess{\`i}, Raileanu, Lomeli,
  Hambro, Zettlemoyer, Cancedda, and Scialom}]{schick2023toolformer}
Schick, T.; Dwivedi-Yu, J.; Dess{\`i}, R.; Raileanu, R.; Lomeli, M.; Hambro,
  E.; Zettlemoyer, L.; Cancedda, N.; and Scialom, T. 2023.
\newblock {Toolformer}: Language Models Can Teach Themselves to Use Tools.
\newblock In \emph{Advances in Neural Information Processing Systems},
  volume~36, 68539--68551.

\bibitem[{Schulman et~al.(2017)Schulman, Wolski, Dhariwal, Radford, and
  Klimov}]{schulman2017ppo}
Schulman, J.; Wolski, F.; Dhariwal, P.; Radford, A.; and Klimov, O. 2017.
\newblock Proximal policy optimization algorithms.
\newblock arXiv:1707.06347.

\bibitem[{Shao et~al.(2024)Shao, Wang, Zhu, Xu, Song, Bi, Zhang, Zhang, Li, Wu
  et~al.}]{shao2024deepseekmath}
Shao, Z.; Wang, P.; Zhu, Q.; Xu, R.; Song, J.; Bi, X.; Zhang, H.; Zhang, M.;
  Li, Y.; Wu, Y.; et~al. 2024.
\newblock {DeepSeekMath}: Pushing the Limits of Mathematical Reasoning in Open
  Language Models.
\newblock arXiv:2402.03300.

\bibitem[{Sheng et~al.(2025)Sheng, Zhang, Ye, Wu, Zhang, Zhang, Peng, Lin, and
  Wu}]{sheng2024verl}
Sheng, G.; Zhang, C.; Ye, Z.; Wu, X.; Zhang, W.; Zhang, R.; Peng, Y.; Lin, H.;
  and Wu, C. 2025.
\newblock {HybridFlow}: A Flexible and Efficient {RLHF} Framework.
\newblock In \emph{Proceedings of the Twentieth European Conference on Computer
  Systems}, EuroSys '25, 1279--1297. New York, NY, USA: Association for
  Computing Machinery.
\newblock ISBN 9798400711961.

\bibitem[{Shridhar et~al.(2021)Shridhar, Yuan, C\^ot\'e, Bisk, Trischler, and
  Hausknecht}]{shridhar2021alfworld}
Shridhar, M.; Yuan, X.; C\^ot\'e, M.-A.; Bisk, Y.; Trischler, A.; and
  Hausknecht, M. 2021.
\newblock {ALFWorld: Aligning Text and Embodied Environments for Interactive
  Learning}.
\newblock In \emph{Proceedings of the International Conference on Learning
  Representations}.

\bibitem[{Tan et~al.(2026)Tan, Yang, Chen, Shao, Wen, Shen, Luo, Du, Guo, and
  Li}]{tan2026hcapo}
Tan, H.-Z.; Yang, X.-W.; Chen, H.; Shao, J.-J.; Wen, Y.; Shen, Y.; Luo, W.; Du,
  X.; Guo, L.-Z.; and Li, Y.-F. 2026.
\newblock Hindsight credit assignment for long-horizon llm agents.
\newblock arXiv:2603.08754.

\bibitem[{Team et~al.(2023)Team, Anil, Borgeaud, Alayrac, Yu, Soricut,
  Schalkwyk, Dai, Hauth, Millican et~al.}]{team2023gemini}
Team, G.; Anil, R.; Borgeaud, S.; Alayrac, J.-B.; Yu, J.; Soricut, R.;
  Schalkwyk, J.; Dai, A.~M.; Hauth, A.; Millican, K.; et~al. 2023.
\newblock Gemini: a family of highly capable multimodal models.
\newblock arXiv:2312.11805.

\bibitem[{Ultes et~al.(2017)Ultes, Barahona, Su, Vandyke, Kim, Casanueva,
  Budzianowski, Mrk{\v{s}}i{\'c}, Wen, Gasic et~al.}]{ultes2017pydial}
Ultes, S.; Barahona, L. M.~R.; Su, P.-H.; Vandyke, D.; Kim, D.; Casanueva, I.;
  Budzianowski, P.; Mrk{\v{s}}i{\'c}, N.; Wen, T.-H.; Gasic, M.; et~al. 2017.
\newblock {PyDial}: A Multi-Domain Statistical Dialogue System Toolkit.
\newblock In \emph{Proceedings of the Annual Meeting of the Association for
  Computational Linguistics, System Demonstrations}, 73--78.

\bibitem[{Wang et~al.(2023)Wang, Wang, Liu, Chen, Yuan, Peng, and
  Ji}]{wang2023mint}
Wang, X.; Wang, Z.; Liu, J.; Chen, Y.; Yuan, L.; Peng, H.; and Ji, H. 2023.
\newblock Mint: Evaluating llms in multi-turn interaction with tools and
  language feedback.
\newblock arXiv:2309.10691.

\bibitem[{Wang et~al.(2025)Wang, Wang, Wang, Zhang, Li, Yang, Jin, Yu, Nguyen,
  Liu et~al.}]{wang2025ragen}
Wang, Z.; Wang, K.; Wang, Q.; Zhang, P.; Li, L.; Yang, Z.; Jin, X.; Yu, K.;
  Nguyen, M.~N.; Liu, L.; et~al. 2025.
\newblock {RAGEN}: Understanding Self-Evolution in {LLM} Agents via Multi-Turn
  Reinforcement Learning.
\newblock arXiv:2504.20073.

\bibitem[{Wei et~al.(2025)Wei, Zeng, Li, Brown, Frunza, Deng, Schneider,
  Nevmyvaka, Zhao, Garcia et~al.}]{zeng2025turnlevel}
Wei, Q.; Zeng, S.; Li, C.; Brown, W.; Frunza, O.; Deng, W.; Schneider, A.;
  Nevmyvaka, Y.; Zhao, Y.~K.; Garcia, A.; et~al. 2025.
\newblock Reinforcing multi-turn reasoning in llm agents via turn-level reward
  design.
\newblock arXiv:2505.11821.

\bibitem[{Xie et~al.(2026)Xie, Thomas, Hansen, Fu, Li, and Wang}]{xie2026tips}
Xie, Y.; Thomas, N.; Hansen, N.; Fu, Y.; Li, L.~E.; and Wang, X. 2026.
\newblock Tips: Turn-level information-potential reward shaping for
  search-augmented llms.
\newblock arXiv:2603.22293.

\bibitem[{Yang et~al.(2025)Yang, Li, Yang, Zhang, Hui, Zheng, Yu, Gao, Huang,
  Lv et~al.}]{yang2025qwen3}
Yang, A.; Li, A.; Yang, B.; Zhang, B.; Hui, B.; Zheng, B.; Yu, B.; Gao, C.;
  Huang, C.; Lv, C.; et~al. 2025.
\newblock Qwen3 Technical Report.
\newblock arXiv:2505.09388.

\bibitem[{Yao et~al.(2022{\natexlab{a}})Yao, Chen, Yang, and
  Narasimhan}]{yao2022webshop}
Yao, S.; Chen, H.; Yang, J.; and Narasimhan, K. 2022{\natexlab{a}}.
\newblock {WebShop}: Towards Scalable Real-World Web Interaction with Grounded
  Language Agents.
\newblock In \emph{Advances in Neural Information Processing Systems},
  volume~35, 20744--20757.

\bibitem[{Yao et~al.(2024)Yao, Shinn, Razavi, and Narasimhan}]{yao2024taubench}
Yao, S.; Shinn, N.; Razavi, P.; and Narasimhan, K. 2024.
\newblock $\tau$-bench: A Benchmark for Tool-Agent-User Interaction in
  Real-World Domains.
\newblock arXiv:2406.12045.

\bibitem[{Yao et~al.(2022{\natexlab{b}})Yao, Zhao, Yu, Du, Shafran, Narasimhan,
  and Cao}]{yao2022react}
Yao, S.; Zhao, J.; Yu, D.; Du, N.; Shafran, I.; Narasimhan, K.; and Cao, Y.
  2022{\natexlab{b}}.
\newblock Re{A}ct: Synergizing Reasoning and Acting in Language Models.
\newblock arXiv:2210.03629.

\bibitem[{Yu et~al.(2026)Yu, Zhang, Zhu, Yuan, Zuo, Yue, Dai, Fan, Liu, Liu
  et~al.}]{yu2026dapo}
Yu, Q.; Zhang, Z.; Zhu, R.; Yuan, Y.; Zuo, X.; Yue, Y.; Dai, W.; Fan, T.; Liu,
  G.; Liu, L.; et~al. 2026.
\newblock Dapo: An open-source llm reinforcement learning system at scale.
\newblock \emph{Advances in Neural Information Processing Systems}, 38:
  113222--113244.

\bibitem[{Zhang et~al.(2026{\natexlab{a}})Zhang, Tang, Li, Han, and
  Tresp}]{zhang2026webarbiter}
Zhang, Y.; Tang, S.; Li, Z.; Han, Z.; and Tresp, V. 2026{\natexlab{a}}.
\newblock {WebArbiter}: A Principle-Guided Reasoning Process Reward Model for
  Web Agents.
\newblock arXiv:2601.21872.

\bibitem[{Zhang et~al.(2026{\natexlab{b}})Zhang, Zhao, Wang, Wang, Liang, Wang,
  Hu, Cao, and Wong}]{zhang2026fission}
Zhang, Z.; Zhao, F.; Wang, R.; Wang, Z.; Liang, B.; Wang, J.; Hu, Y.; Cao, S.;
  and Wong, K.-F. 2026{\natexlab{b}}.
\newblock Robust Tool Use via Fission-{GRPO}: Learning to Recover from
  Execution Errors.
\newblock In \emph{Proceedings of the Conference of the Association for
  Computational Linguistics}.

\bibitem[{Zhao et~al.(2025)Zhao, Wang, Ma, Kong, Yang, Tuo, Shi, Zhai, and
  Cai}]{zhao2025muarl}
Zhao, W.; Wang, X.; Ma, C.; Kong, L.; Yang, Z.; Tuo, M.; Shi, X.; Zhai, Y.; and
  Cai, X. 2025.
\newblock {MUA-RL}: Multi-Turn User-Interacting Agent Reinforcement Learning
  for Agentic Tool Use.
\newblock arXiv:2508.18669.

\bibitem[{Zhao et~al.(2023)Zhao, Gu, Varma, Luo, Huang, Xu, Wright,
  Shojanazeri, Ott, Shleifer et~al.}]{zhao2023pytorch}
Zhao, Y.; Gu, A.; Varma, R.; Luo, L.; Huang, C.-C.; Xu, M.; Wright, L.;
  Shojanazeri, H.; Ott, M.; Shleifer, S.; et~al. 2023.
\newblock {PyTorch} {FSDP}: Experiences on Scaling Fully Sharded Data Parallel.
\newblock arXiv:2304.11277.

\bibitem[{Zheng et~al.(2025)Zheng, Liu, Li, Chen, Yu, Gao, Dang, Liu, Men, Yang
  et~al.}]{zheng2025gspo}
Zheng, C.; Liu, S.; Li, M.; Chen, X.-H.; Yu, B.; Gao, C.; Dang, K.; Liu, Y.;
  Men, R.; Yang, A.; et~al. 2025.
\newblock Group sequence policy optimization.
\newblock arXiv:2507.18071.

\bibitem[{Zhou et~al.(2024{\natexlab{a}})Zhou, Xu, Zhu, Zhou, Lo, Sridhar,
  Cheng, Ou, Bisk, Fried et~al.}]{zhou2024webarena}
Zhou, S.; Xu, F.~F.; Zhu, H.; Zhou, X.; Lo, R.; Sridhar, A.; Cheng, X.; Ou, T.;
  Bisk, Y.; Fried, D.; et~al. 2024{\natexlab{a}}.
\newblock Webarena: A realistic web environment for building autonomous agents.
\newblock In \emph{International Conference on Learning Representations},
  volume 2024, 15585--15606.

\bibitem[{Zhou et~al.(2025)Zhou, Jiang, Tian, Weston, Levine, Sukhbaatar, and
  Li}]{zhou2025sweetrl}
Zhou, Y.; Jiang, S.; Tian, Y.; Weston, J.; Levine, S.; Sukhbaatar, S.; and Li,
  X. 2025.
\newblock {SWEET-RL}: Training Multi-Turn {LLM} Agents on Collaborative
  Reasoning Tasks.
\newblock arXiv:2503.15478.

\bibitem[{Zhou et~al.(2024{\natexlab{b}})Zhou, Zanette, Pan, Levine, and
  Kumar}]{zhou2024archer}
Zhou, Y.; Zanette, A.; Pan, J.; Levine, S.; and Kumar, A. 2024{\natexlab{b}}.
\newblock Archer: Training language model agents via hierarchical multi-turn
  rl.
\newblock arXiv:2402.19446.

\end{thebibliography}
\end{document}